\documentclass[preprint,3p,times]{elsarticle}
\usepackage{amsmath,amssymb}
\usepackage{booktabs}
\usepackage{longtable}
\usepackage{array}
\usepackage{tabularx}
\usepackage{threeparttable}
\usepackage{siunitx}
\usepackage{graphicx}
\usepackage{xcolor}
\usepackage{tikz}
\usetikzlibrary{arrows.meta,decorations.pathreplacing,positioning}

\newcolumntype{Y}{>{\raggedright\arraybackslash}X}
\newcommand{\rev}[1]{{\color{black}#1}}

\usepackage{lineno}
\modulolinenumbers[2]

\usepackage{hyperref}
\hypersetup{hidelinks}
\journal{TrAC Trends in Analytical Chemistry}

\biboptions{sort&compress}

\begin{document}

% Highlights are required by TrAC at submission and should be uploaded as a
% separate editable file. Do not include them in the main manuscript PDF.
%\input{highlights.tex}

\begin{frontmatter}

\title{Convolutional Neural Networks in Vis-NIR Chemometrics: From Contradiction to Conditional Design}

\author[aff1]{Dário Passos\corref{cor1}}
\ead{dmpassos@ualg.pt}
\cortext[cor1]{Corresponding author}

\affiliation[aff1]{organization={DeepLight Laboratory, Departamento de Física, Faculdade de Ciências e Tecnologia da Universidade do Algarve}, city={Faro}, postcode={8005-189}, country={Portugal}}

\begin{abstract}
\rev{
Near-infrared spectroscopy (NIR and Vis--NIR) is widely used for rapid, non-destructive analysis in food, agriculture, pharmaceutical, process analytical technology (PAT), and bioprocess-monitoring applications. Yet, deep-learning studies in NIR chemometrics often report conflicting conclusions about convolutional neural network (CNN) design: small versus large kernels, shallow versus deep architectures, raw spectra versus preprocessing, compact models versus multi-scale networks, and random-split performance versus transfer robustness. This review argues that many of these apparent contradictions arise from incomplete conditioning rather than from inherently incompatible results. CNN performance in NIR chemometrics depends on the interaction between spectral physics, dataset regime, acquisition protocol, validation design, and deployment scenario. We therefore organize the literature around three central moderators. First, NIR signals are indirect, highly collinear, and often dominated by broad overlapping bands, scattering, temperature, and matrix effects. Second, convolutional design choices should be interpreted through receptive-field reasoning: kernel size, depth, dilation, and multi-scale branches determine which wavelength spans are available to the model, whereas the effective receptive field determines which parts of that span are actually used. Third, validation design can act as a hidden hyperparameter, because random splits may favour architectures that exploit shared batch, instrument, season, or process-run structure rather than transferable chemical information. Building on these points, we propose a conditional design framework in which preprocessing, architecture, hyperparameter tuning, transfer evaluation, interpretability, and reproducibility are treated as coupled components of the modelling pipeline. The goal is not to identify a universally optimal CNN for NIR spectra, but to move CNN--NIR chemometrics toward physics-aware, shift-aware, and reproducible model comparison.}
\end{abstract}


\end{frontmatter}

%\linenumbers

\section{Introduction}
Near-infrared spectroscopy \rev{(NIRS), spanning roughly from 750 to 2500~nm}, is arguably one of the most practical analytical tools for rapid, non-destructive quality assessment across food, agriculture, pharmaceutical, \rev{process analytical technology (PAT) and bioprocess-monitoring} applications \citep{cen2007, Roggo2007_NIR_Chemometrics_Pharma, yeung1999bioprocess, walsh2020point, Pu2020_NIR_PAT_Dairy, anderson2022fruitreview}. Its utility stems from the fact that overtone and combination bands of fundamental molecular vibrations (mainly O--H, C--H, and N--H stretching modes) can be interrogated with minimal sample preparation, thereby enabling high-throughput screening and (increasingly) inline monitoring. However, the same spectral physics that makes NIR broadly applicable also makes the resulting data structure unusually challenging. Absorption bands are broad and heavily overlapping, wavelength collinearity is \rev{high}, and spectra can be strongly modulated by physical effects (e.g., temperature, scattering geometry, path-length variability, and instrument response) \citep{cen2007,bechuck2019simulation}. \rev{Given the broad application scope of this technique and the persistent difficulty of extracting robust chemical information from such data, it was just a matter of time until the Chemometrics community leaned into Deep-Learning (DL) as a prospective modelling tool. However, DL model design in NIR chemometrics is not directly analogous to the usual DL benchmarks on which most modern architectures were developed and validated. The use of DL does not remove the classical constraints of NIR chemometrics (comparatively small datasets, strong wavelength-to-wavelength collinearity, and domain shifts due to instrument, temperature, scattering, packaging, etc.) that can dominate the error budget, rather, it transfers them into the problem of architecture design. This makes NIR chemometric modelling only partially comparable to large, label-rich, and comparatively standardized domains such as computer vision, from which many contemporary DL design habits were inherited.}

\rev{From the several currently available neural network architectures}, one-dimensional convolutional neural networks (1D-CNNs) have attracted considerable attention for spectral regression and classification, with multiple studies reporting competitive (and often superior) performance relative to classical chemometric baselines such as partial least squares (PLS) regression \citep{malek2018onecnn,acquarelli2017cnn,cuifearn2018cnn,zhang2022mlreview, walsh2023cnnfruitreview, luo2024cnnreview}. \rev{Table \ref{tab:nir_cnn_literature} presents some recent representative works on this topic.} Nevertheless, the literature exhibits a recurring and practically problematic pattern: mutually incompatible conclusions regarding nearly every major design decision. Some studies favor compact, single-convolution or shallow-kernel architectures \citep{malek2018onecnn,cuifearn2018cnn,chenwang2019e2e}, whereas others report that multi-scale, residual, or dilated designs are necessary to obtain robust performance (particularly under shift) \citep{zhang2019deepspectra,gan2023dilated,martins2022spectranet53,tan2023inceptionresnet}. Similarly, some authors argue that end-to-end raw spectral input can render preprocessing obsolete \citep{chenwang2019e2e,zhang2019deepspectra}, while others demonstrate clear gains from combining preprocessing with deep learning \citep{helin2021preprocessing,mishra2021synergy,walsh2024mango}. The net effect is a field in which architectural choices are often justified \textit{post hoc} (by selective citation), and in which practitioners face a confusing landscape of apparently contradictory recommendations \citep{MishraTrACReview2022}.

%%REV: Tabela adicionada em resposta ao reviewer 1.

\begin{table*}[htbp]
	\centering
	\caption{\rev{Compact cross-domain overview of CNN-based NIR/Vis--NIR chemometrics applications in the recent literature}}
	\label{tab:nir_cnn_literature}
	\small
	%\color{red}
	\begin{tabularx}{\textwidth}{p{0.2\textwidth}p{0.1\textwidth}Y}
		\toprule
		\textbf{Topic} & \textbf{References} & \textbf{General description} \\
		\midrule
		Fruit internal quality and postharvest control
		& \citep{mishra2022multioutput,martins2022spectranet53,Escarate2022StoneFruit,walsh2024mango, walsh2023cnnfruitreview}
		& Intact-fruit quality assessment using NIR/Vis--NIR spectra: multi-output prediction of physicochemical traits, soluble-solids content in oranges and stone fruit, cultivar/species discrimination, and mango dry-matter estimation with 1D-CNN architectures. \\
		\addlinespace
		Food authentication, safety, and processed-food control
		& \citep{NallanChakravartula2022CoffeeCNN,Mansuri2022MaizeFungalCNN,Wang2022AflatoxinMTFCNN,Sun2023PastaCNNLSTM}
		& CNN-based spectral models for quantifying coffee adulteration, detecting fungal contamination in maize kernels, monitoring aflatoxin B$_1$ in maize from NIR spectra, and classifying pre-cooked pasta products under different physical states. \\
		\addlinespace
		Circular economy and material sorting
		& \citep{Xia2021PlasticCNN,Maliks2021PlasticBottleCNN,Du2022WasteTextileCNN,Riba2022TextileWasteCNN}
		& NIR and multispectral-NIR spectra combined with CNNs for polymer and recyclable-material identification, including black-plastic discrimination, plastic-bottle sorting, online waste-textile sorting, and classification of pure and blended post-consumer textile fibres. \\
		\addlinespace
		Soil spectroscopy and agronomy
		& \citep{Kawamura2021SoilP1DCNN,HosseinpourZarnaq2023SoilCNN,Miao2024SoilSOMLSTMCNN}
		& 1D and hybrid CNN models applied to Vis--NIR/NIR soil spectra and regional spectral libraries for available phosphorus, soil physicochemical properties, and soil organic matter prediction, often benchmarked against PLS, random forests, and local regression methods. \\
		\addlinespace
		Wood and forestry authentication
		& \citep{Pan2023SoftwoodCNN,Pan2024WoodPortableCNN,Li2025WoodMultispectralCNN}
		& CNN and related deep-learning models applied to NIR or multispectral wood spectra for non-destructive wood-species identification, including raw laboratory NIR spectra, portable short-wave NIR spectrometers, and comparison of NIR, hyperspectral, and terahertz spectral modalities for wood authentication. \\
		\addlinespace
		Pharmaceutical analysis, PAT, and authenticity screening
		& \citep{Peng2024GranulationCBAMCNN,Han2025NIRTransferLearning,Awotunde2022FalsifiedPharmaNIR}
		& Deep-learning and machine-learning NIR workflows for pharmaceutical process monitoring and screening: particle-size distribution in fluid-bed granulation, transfer learning for NIR model adaptation, and discrimination of substandard/falsified formulations from genuine products. \\
		\addlinespace
		Industrial bioprocessing and fermentation
		& \citep{Banerjee2024FermentationICNN,Gangwar2025CellCultureCNN,Zheng2024OolongFermentationCNN}
		& CNN-enabled NIR/Vis--NIR chemometrics for real-time or at-line process monitoring: multianalyte quantification in microbial fermentation, cell-culture media characterization for bioprocess control, and online monitoring of tea fermentation degree using Vis--NIR/image data fusion. \\
		\addlinespace
		Biomedical and physiological NIRS
		& \citep{Kwon2021FNIRSBCI,Park2024TCNNBreathing}
		& CNN models for physiological NIRS signals, including subject-independent functional-NIRS brain--computer interfaces and wearable NIRS-based breathing-pattern classification for respiratory monitoring. \\
		\bottomrule
	\end{tabularx}
\end{table*}

\rev{Although DL applications in NIR chemometrics have grown rapidly, the evidence base remains fragmented across application domains and venues. Recent reviews document this expansion across food analysis, general NIR spectroscopy, and AI-enabled spectroscopy, but they also show that papers differ substantially in reporting of sample size, spectral range/resolution, preprocessing, tuning strategy, split design, and external validation \citep{zhang2022mlreview,luo2024cnnreview,jia2024aidriven}. In contrast to areas of machine learning where a small number of research communities iterate systematically on architectures, objectives, and benchmarks, much of the NIR DL literature is application-led. Studies often adapt an existing model to a specific dataset and emphasize predictive outcomes, while providing limited analysis of architectural mechanisms, inductive biases, protocol sensitivity, or data/code availability. A further practical consequence is diffusion of the evidence base across heterogeneous venue, ranging from core chemometrics journals (e.g., \textit{Chemometrics and Intelligent Laboratory Systems} and \textit{Journal of Chemometrics}) to analytical chemistry, food-quality, instrumentation, petroleum, pharmaceutical, and engineering journals. This makes it difficult to track, compare, and cumulatively interpret what constitutes genuine methodological progress towards NIR-CNN architecture design.}

We argue that these situations are neither accidental nor solely a symptom of field immaturity. \rev{Rather, they are a predictable outcome of interacting factors:} (1)~the physics of indirect measurement in NIR systems, which often yields signals whose informative structure is broad, confound-laden, and shaped by matrix and acquisition effects (as opposed to isolated analyte peaks); (2)~the relationship between convolutional kernel size, network depth, and receptive field, which determines the scale of spectral structure a model can represent; and (3)~validation design (split strategy, hyperparameter tuning budget, preprocessing choices, data availability and explicit external validation) acting as a hidden but powerful determinant of model \rev{performance}. When these moderators are uncontrolled across studies, conflicting findings are not merely possible, they are structurally expected.

With this short review we intend to clear the current impasse by:
\begin{enumerate}
\item Explaining why contradictory CNN hyperparameter findings are expected in NIR (and Vis-NIR) contexts, by tracing them to the interaction of spectral physics, receptive-field mechanics, and protocol design.
\item \rev{Proposing the adoption of conditional design rules that link architecture choices to measurable properties of the spectral data and deployment scenario, while explicitly avoiding the claim that a universal CNN recipe for spectral data is currently available.}
\end{enumerate}

\rev{The current review focuses on CNN architecture design for NIR chemometrics, while also mentioning other DL (trendy) models such as Transformer and hybrid CNN--Transformer models where they illuminate the same design problem. Throughout the manuscript we will often make use of fruit spectra and associated chemometric tasks to illustrate some points. These spectra are used as illustrative anchors because they expose (broad) water-bands, temperature, and scattering confounds clearly. They are not intended to imply that the framework is restricted to fruit or that the same numerical choices transfer unchanged to other chemical matrices.} We prioritize studies that report explicit protocol details (split strategy, preprocessing pipeline, hyperparameter search space, acquisition protocol, data/code availability, and external validation) because disagreement in this field appears to be driven at least as much by protocol as by architecture \citep{dirks2022hpo,passos2022tutorial,jia2024aidriven}. Works that provide only in-domain random-split accuracy without split rationale, seed reporting, uncertainty quantification, or reuse-enabling data/code information are noted for context but are assigned lower evidential weight.

\section{The Physics of Indirect Measurement in Vis-NIR Spectroscopy}

To understand why CNN-design contradictions arise in NIR chemometrics, it is useful to start from the physical nature of the signal these models attempt to learn. Unlike mid-infrared spectroscopy, where fundamental vibrational modes often yield relatively sharp and well-separated absorption bands, the near-infrared region is dominated by overtone and combination bands that are inherently broad, weak, and extensively overlapping. \rev{To make the argument concrete, we use intact fruit quality assessment as a recurring example; the same reasoning extends to other matrix-sensitive NIR applications, but the relevant wavelength ranges, feature widths, and confounds must be re-estimated for each matrix and acquisition protocol. In the short-wave NIR (SW-NIR) or Vis--NIR window of approximately 500--1200~nm (widely used in fruit spectroscopy) the spectral landscape is typically dominated by the second and third overtones of O--H stretching in water, centered near 970~nm with a full width at half maximum (FWHM) of roughly 50--100~nm (depending on temperature and matrix composition) \citep{cen2007,walsh2020point,bechuck2019simulation}.}

A typical example is soluble solids content (SSC, often reported as Brix) in fresh fruit. SSC is largely composed of sugars (sucrose, glucose, and fructose), and the C--H overtone signatures of these carbohydrates do appear in the Vis-NIR region, most notably as third and fourth overtones near 910--930~nm, with weaker contributions around 850--870~nm. Yet, they are typically orders of magnitude weaker than the water signal. In high-moisture matrices (often 85--90\% water), these sugar-associated contributions can be effectively buried within the shoulders and flanks of the dominant water envelope and may approach the instrument noise floor \citep{walsh2020point,anderson2022fruitreview}. Therefore, a central credibility question follows naturally: if a ``sugar peak'' is not cleanly resolvable from the water envelope, what is the model actually measuring when it reports high predictive accuracy (e.g., $R^2>0.9$) for SSC?

We argue that the answer is often not analyte-centric spectroscopy (i.e., ``the model directly measures sugar absorption''), but rather matrix-centric spectroscopy, where the analyte modulates a water-dominated background in systematic ways. In this regard, it is useful to think using the framework of \textit{aquaphotomics}, developed by Tsenkova and colleagues. It formalizes the principle that water in biological systems is not a single spectral species but a dynamic equilibrium of hydrogen-bonded molecular assemblies \citep{muncan2019aquaphotomics,roger2022aquapre}. Different structural populations (e.g., weakly hydrogen-bonded (``free'') water, small clusters, and tetrahedrally coordinated networks) contribute to the composite absorption envelope at slightly shifted wavelengths. Changes in solute concentration, temperature, or microstructure can perturb this equilibrium, thereby deforming broad water bands in a manner that is reproducible (and, crucially, learnable). \rev{This overall reasoning can be adapted to other types of chemical matrices and chemometric tasks that can be prone to different types of interferants.}

In practical terms, the informative signal for SSC prediction may be a subtle change in the \textit{shape} of the 970~nm water band rather than an isolated carbohydrate peak. For instance a small center shift, an asymmetric deformation of the flanks or a modest change in full width at half maximum (FWHM), i.e., a measure of peak broadness. Hence, what appears as ``chemical prediction'' can in fact be mediated by water-structure dynamics and scattering effects that co-vary with the target trait. This is not a weakness of NIR spectroscopy; it is its defining character in water-rich matrices. However, it does imply that model design (and, especially, validation design) must be aligned with this indirect-measurement regime.

\rev{A second implication is that NIR models are vulnerable to confounds that act through the same water-dominated channels. Temperature shifts can move and reshape water bands, alter hydrogen-bond equilibria, and change apparent analyte relationships \citep{roger2003epo,chauchard2004temperature,hageman2005temperature,liu2017temperature,sun2020temp}. This is not a minor nuisance variable: temperature compensation has been treated as a calibration-transfer problem in classical chemometrics, and recent studies show that temperature can itself be predicted from Vis--NIR spectra \citep{martins2023spectranet32}. Scattering changes (e.g., from particle size, texture, packaging, probe contact, path length, or surface geometry) can also induce multiplicative and baseline effects that are strongly correlated within acquisition batches \citep{luo2018local}. Therefore, strong performance within a given dataset should not be interpreted, by itself, as evidence that a model has learned a chemically transferable relationship. In NIR data, a CNN may instead exploit correlations that are stable within the calibration domain, such as instrument-specific baselines, scattering patterns, temperature-dependent variation, packaging effects, or sample-presentation artefacts, but that break down when the acquisition conditions change. This distinction is central to the interpretation of CNN-based NIR studies. Architectural choices such as kernel size, receptive field, and multi-scale processing determine which spectral structures the model can preferentially exploit, while acquisition and evaluation protocols determine whether those structures are genuinely robust. For this reason, conclusions about CNN performance in NIR chemometrics depend not only on the network architecture, but also on the use of appropriate validation strategies.}

\section{Convolution, Receptive Fields, and the Kernel Size Debate}

With the physical signal structure established, we can now examine how 1D-CNNs interact with NIR spectra and, consequently, why the kernel-size debate (small versus large) has proven so resistant to consensus.

A one-dimensional convolution applied to a spectral input vector $x$ with a kernel $w$ of size $k$ computes, at each position $i$, the sliding inner product:
\begin{equation}
y[i] = \sum_{j=0}^{k-1} w[j] \cdot x[i+j].
\end{equation}

The kernel size $k$ defines the window (i.e., the number of consecutive wavelength channels) that the unit (or neuron) processes at each step. This choice is not a neutral architectural detail because it sets a strong inductive bias regarding the \textit{scale} and \textit{type} of spectral structure the network can represent efficiently. In deep learning, a convolution typically refers to a learned cross-correlation (the kernel is not reversed) whose filter coefficients are optimized from data \citep{Scardapane2024Book}, whereas in classical signal processing convolution is a fixed, predefined linear operation with kernel reversal, usually chosen to implement a specific physical/analytical filter.

Small kernels (typically $k=3$ or $5$) behave as local feature detectors. In the spectral domain, a trained $k=3$ kernel can approximate a first-difference (local slope) or second-difference (local curvature) operator, effectively learning a derivative-like preprocessing step. Indeed, several studies report that first-layer CNN filters trained on raw spectra converge toward derivative-like or smoothing operations \citep{cuifearn2018cnn,helin2021preprocessing,Zhang2020UnderstandingCNNs,passos2025deeptuttifrutti2}, thereby re-discovering (in a data-driven manner) what chemometricians have long implemented explicitly (e.g., Savitzky--Golay derivatives). This connection is not incidental: derivative preprocessing suppresses baseline offset and low-frequency drift, and a CNN that learns a similar transform is responding to the same optimization pressures.

Large kernels (e.g., $k=11,21,31+$) serve a different role. A single large kernel can span a substantial fraction of a broad spectral feature and can therefore operate as a waveform pattern matcher. It can encode the curvature of a shoulder, the asymmetry of an envelope, or the overall ``shape'' of a band. Where a small kernel often perceives only a near-linear segment (and may struggle to distinguish the rising edge of an absorption band from a slow baseline drift), a large kernel can incorporate wider context and discriminate shape from trend.

This distinction maps directly onto the physics described in the preceding section. If the informative signal for SSC prediction is a subtle deformation of the 970~nm water band, for example a small shift in band centre or a 5--15\% change in FWHM, then the model must compare intensities across a \rev{relatively wide wavelength interval. At typical wavelength sampling intervals of 1--3~nm per channel, a first-layer $k=3$ convolution initially aggregates information from only about 3--9~nm of the spectrum, i.e., the local receptive field of one unit. This is only a narrow slice of a feature that may span 50--100~nm, over which the band can appear almost linear. A first-layer $k=31$ convolution}, in contrast, spans roughly 30--90~nm and can directly encode whether the local envelope is broad or narrow, symmetric or skewed. However, as information propagates towards deeper convolutional layers, even small kernel features can aggregate into larger ones.

\begin{figure}[t]
  \centering
  \includegraphics[width=0.5\linewidth]{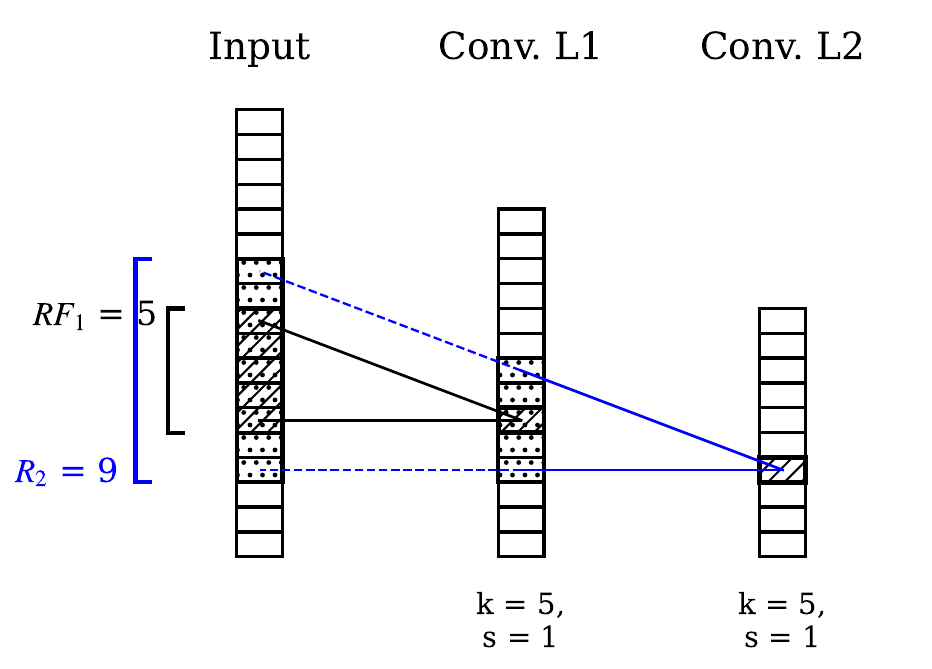}
  \caption{Example of receptive field (RF) growth in a 1D-CNN with kernel size $k=5$ and stride $s=1$ across two convolutional layers. While a unit in the first convolutional layers "sees" 5 input features, an individual units in the second convolutional layer "sees" a wider portion of the input vector (spectrum).}
  \label{fig:rf_1dcnn}
\end{figure}

\rev{This is the architectural lesson associated with VGG-style image networks, named after the Oxford Visual Geometry Group, where several small $3 \times 3$ convolutions were stacked to approximate the receptive field of larger filters while introducing additional nonlinearities \citep{simonyan2014vgg, Scardapane2024Book}. The same principle applies in one-dimensional spectral CNNs. In the example shown in Fig.~\ref{fig:rf_1dcnn}, the left layer represents the input spectrum, whereas Conv.~L1 and Conv.~L2 are two successive convolutional layers. With $k=5$, stride $s=1$, and dilation $d=1$, one unit in Conv.~L1 has a theoretical receptive field of five input wavelength channels. One unit in Conv.~L2 has a receptive field of nine input channels, because it combines five neighbouring Conv.~L1 activations with overlapping input supports. However, in NIR chemometrics this should be read only as an architectural analogy. The relevant question is not whether stacking can expand the theoretical receptive field, but whether the \textit{effective} receptive field (ERF) learned from small, highly collinear datasets covers the chemically informative wavelength scale and remains robust under realistic sources of spectral shift}. In general, the theoretical RF after $L$ convolutional layers can be written as
\begin{equation}
\mathrm{RF}_{L} = 1+ \sum_{i=1}^{L} (k_i-1)~d_i
\left( \prod_{j=1}^{i-1} s_j \right),
\end{equation}
where $k_i$, $d_i$, and $s_i$ denote the kernel size, dilation factor, and stride of layer $i$, respectively, and the product term is taken as one for $i=1$.

\begin{figure}[t]
  \centering
  \resizebox{0.98\linewidth}{!}{%
  \begin{tikzpicture}[
    font=\scriptsize,
    >=Latex,
    every node/.style={text=black},
    redline/.style={draw=red,line width=0.7pt},
    blackline/.style={draw=black,line width=0.7pt},
    blueline/.style={draw=blue,line width=0.7pt},
    redarrow/.style={->,draw=black,line width=0.7pt},
    redbrace/.style={decorate,decoration={brace,amplitude=4pt},draw=black,line width=0.7pt},
    redbox/.style={draw=red,rounded corners=1pt,align=center,inner sep=3pt,text=black},
    box/.style={draw=black,rounded corners=1pt,align=center,inner sep=3pt,text=black},
    smallbox/.style={draw=black,minimum width=0.36cm,minimum height=0.26cm,inner sep=0pt}
  ]
    \fill[blue!5] (1.20,0) rectangle (7.20,2.58);
    \draw[blackline,->] (0,0) -- (12.35,0) node[below right] {wavelength};
    \foreach \x/\lab in {0/900,3/950,6/1000,9/1050,12/1100}
      \draw[blackline] (\x,0.07) -- (\x,-0.07) node[below=3pt] {\lab~nm};
    %\draw[redline,dotted] (4.20,0) -- (4.20,2.28);
    %\node[anchor=north] at (4.20,-0.12) {970~nm};
    \draw[blueline,thick,smooth] plot coordinates {(0.2,0.20) (1.1,0.42) (2.0,0.82) (3.0,1.55) (3.6,2.02) (4.2,2.24) (4.8,2.02) (5.6,1.55) (6.6,0.82) (7.7,0.42) (9.2,0.24) (11.7,0.18)};
    \draw[redbrace] (1.20,2.72) -- (7.20,2.72) node[midway,above=5pt] {};
    \node[align=center,anchor=south] at (4.20,2.99) {Broad peak spanning tens of channels};

    \draw[blackline,dashed] (2.95,0.08) rectangle (3.47,1.18);
    % \node[align=center,anchor=north] at (3.21,-0.42) {small kernel\\local slope};
    \draw[blackline, dashed] (1.65,0.16) rectangle (6.75,1.70);
    \node[align=center,anchor=south] at (4.20,1.65) {theoretical RF};
    \draw[redline,very thick,smooth] plot coordinates {(2.10,1.55) (2.55,1.62) (3.35,2.02) (4.20,2.24) (5.05,2.02) (5.85,1.62) (6.20,1.55)};
    \node[align=center,anchor=south, text=red] at (4.15,2.25) {ERF concentrated near center};

    \node[box,minimum width=2.75cm,minimum height=0.78cm] (compact) at (1.75,-1.72) {small kernel \\$k=3$--$5$};
    \foreach \x in {0.92,1.32,1.72,2.12,2.52}
      \node[smallbox] at (\x,-2.38) {};
    \draw[redarrow] (compact.north) -- (3.20,0.10);

    \node[box,minimum width=2.95cm,minimum height=0.78cm] (dilated) at (6.00,-1.72) {large RF/ERF\\large $k$, depth, dilation};
    \foreach \x in {5.05,5.55,6.05,6.55,7.05}
      \node[smallbox] at (\x,-2.38) {};
    \draw[blackline] (5.05,-2.38) -- (7.05,-2.38);
    \draw[redarrow] (dilated.north) -- (4.20,0.18);

    \node[box,minimum width=2.95cm,minimum height=0.78cm] (multi) at (10.20,-1.72) {multi-scale branches\\mixed feature widths};
    \draw[blackline] (9.25,-2.20) -- (11.15,-2.20);
    \draw[blackline] (9.45,-2.42) -- (10.95,-2.42);
    \draw[blackline] (9.70,-2.64) -- (10.70,-2.64);
    \draw[redarrow] (multi.north) -- (6.65,0.18);
   % \node[box,fill=red!4,minimum width=7.85cm,minimum height=0.72cm] at (6.00,-3.38) {choose/search architectures whose ERF matches feature width, then validate under deployment-like shift};
  \end{tikzpicture}}
  \caption{\rev{
Schematic relation between spectral feature width, theoretical receptive field (RF), effective receptive field (ERF), and CNN design choices in one-dimensional Vis--NIR spectra. A broad absorption feature, illustrated here as a water-related band (in blue) centred near 970~nm, may span tens of wavelength channels. A small convolutional kernel ($k=3$--5) samples only a local portion of the band, such as a shoulder or local slope, whereas a larger theoretical RF provides access to a wider spectral interval. However, the input region that materially contributes to the model output may be more concentrated than the theoretical RF, as indicated by the ERF near the band centre (in red). Broad or mixed-scale spectral structures therefore motivate design choices such as larger kernels, sufficient depth, dilation, or multi-scale branches that combine narrow and broad spectral features.
}}
  \label{fig:rf_1dcnn2}
\end{figure}
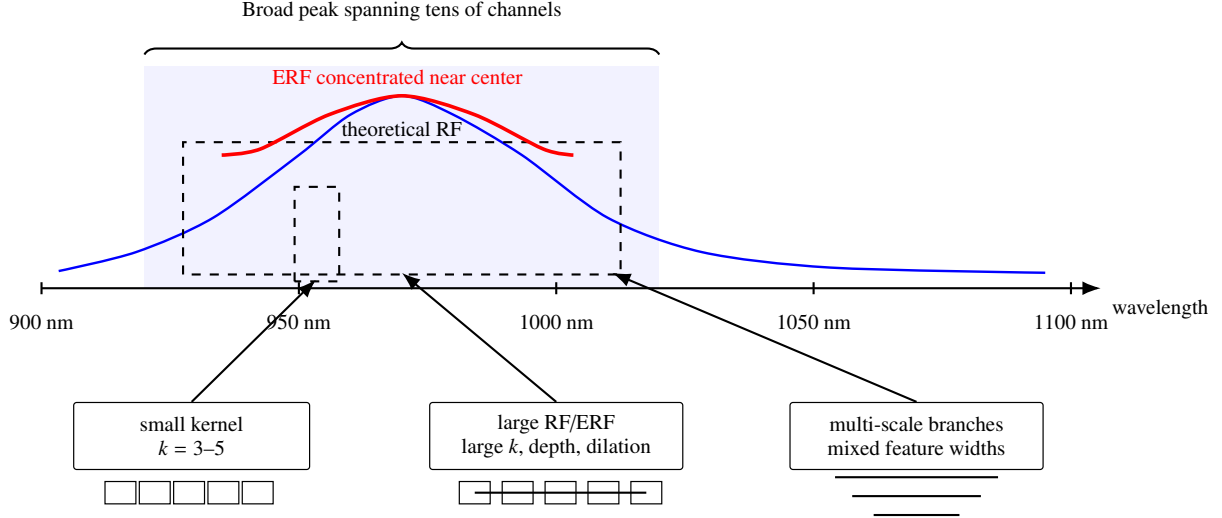

\rev{The theoretical RF is therefore a useful first-order design quantity for spectral CNNs, because it links an architectural choice, such as kernel size, depth, stride, or dilation, to a physically interpretable wavelength span. In NIR chemometrics, this provides a more principled starting point than treating kernel width as a purely empirical hyperparameter: the accessible spectral context can be compared with the expected width of chemically informative bands, shoulders, or matrix-induced envelope deformations. However, this interpretation should be used as a design prior rather than as a guarantee of what the trained network will actually exploit. The theoretical RF defines the maximum input region that can influence a given activation, but the effective receptive field (ERF) describes the subset of that region that materially contributes to the output. For CNNs applied to images, \citet{luo2017erf} showed that the ERF is typically smaller than the theoretical RF and often exhibits a Gaussian-like concentration around the centre. Intuitively, this occurs because central input positions are connected to a deep unit (or activation) through more short computational paths than peripheral positions, causing their gradients to accumulate more strongly. Consequently, the edges of the theoretical RF contribute exponentially less to the learned representation. An analogous caution applies to one-dimensional chemometric CNNs: a small-kernel CNN may have a nominal RF (in the deeper layers) that reaches part of a broad absorption band, while its ERF may still be concentrated on a local shoulder or slope segment. This distinction matters for Vis--NIR spectra because water bands and matrix-induced envelope changes can span tens of wavelength channels. Thus, RF-based reasoning can guide kernel-size, depth, dilation, and multi-scale design, but the resulting architecture should still be interpreted and validated with awareness that the material spectral evidence used by the model may be narrower than the theoretical RF. Increasing depth can expand the theoretical RF and, often, the ERF, but in small chemometric datasets this must be balanced against optimization difficulty, overfitting risk, regularization, tuning budget, and validation design. See Fig.~\ref{fig:rf_1dcnn2} for a visual example.}

\rev{This observation clarifies why the ``small versus large kernel'' debate is often confounded. What is often described simply as a change in kernel size also changes, or interacts with, the theoretical RF, the ERF, and the total parameter count. Dilated convolutions illustrate this point particularly well. By inserting gaps between kernel elements, dilation expands the theoretical RF, often rapidly with depth, while preserving the parameter efficiency of small kernels \citep{yu2016dilated,gan2023dilated}. In Vis--NIR settings, this can be interpreted as a mechanism for comparing intensities at wavelength separations commensurate with broad feature widths, i.e., implementing derivative-like or contrast-like operations at a controlled spectral scale.}

\rev{To isolate the geometry of the argument, consider one input and one output channel and ignore bias terms. A compact CNN with $k=3$ and three non-dilated layers has a theoretical RF of 7, but potentially a substantially smaller ERF, with 9 kernel coefficients across the three layers in this simplified single-channel case. More generally, parameter count scales with the number of input and output channels at each layer. A single $k=31$ layer has an accessible span of 31 wavelength channels and 31 coefficients per input--output channel pair, although the learned weights still determine which parts of this span are actually used. A three-layer $k=3$ network with dilation factors 1, 2, and 4 achieves a theoretical RF of 15 with the same single-channel coefficient count as the non-dilated three-layer example. These are therefore not merely different ``kernel sizes''; they are different inductive biases under different parameterizations.}

\rev{Consequently, when studies compare ``small'' and ``large'' kernels without matching, or at least explicitly reporting, theoretical RF, ERF-related behaviour, channel width, total parameter count, optimizer, regularization, and validation protocol, the resulting performance differences are difficult to interpret. This caution is especially important in small-dataset regimes, where parameter efficiency and regularization often matter as much as architectural expressivity \citep{cuifearn2018cnn,dirks2022hpo,passos2022tutorial}.}

Multi-scale architectures offer a complementary strategy. Inception-style modules \citep{szegedy2016inception} process the input through parallel branches with different kernel sizes and concatenate the resulting representations, allowing subsequent layers to learn how much weight to assign to each scale. This is particularly natural for spectra where narrow features (e.g., pigment-related bands in the visible) coexist with broad envelopes. DeepSpectra \citep{zhang2019deepspectra}, IPA \citep{haffner2025ipa} and subsequent multi-scale NIR architectures \citep{yu2023multiscale,tan2023inceptionresnet,martins2022spectranet53,martins2023spectranet32} follow this logic: large-kernel branches can capture broad band deformations while small-kernel branches capture local edges or narrow bands, and the model can learn scale relevance in a task-dependent manner. \rev{Recent bioprocessing applications extend the same idea beyond food and agricultural matrices, showing that multi-branch CNNs can extract multi-analyte prediction features from complex microbial fermentation systems directly from NIR spectra \citep{Banerjee2024FermentationICNN}.}

Ultimately, the kernel-size debate is best reframed as a question of \textit{ERF relative to informative feature width}. Once this framing is adopted, apparently contradictory literature results become expected rather than surprising: studies focused on narrow spectral windows, sharp features, or classification tasks dominated by local discriminative cues may favor small-kernel designs; studies involving broad bands, regression on latent physical traits, and water-dominated spectra tend to favor large ERFs, whether achieved via large kernels, dilation, depth, or multi-scale processing.

\section{Validation Design as a Hidden Hyperparameter}

\rev{Even if architectural comparisons were to rigorously control ERF and parameter count, a second major source of contradiction would remain, the \textit{validation protocol}. In many CNN--NIR studies, the choice of how data are split into training, validation, and test sets receives far less attention than network architecture. Nevertheless, its impact on reported performance, and on model ranking, can rival or exceed that of many individual architectural changes, depending on the magnitude of the deployment shift.}

\rev{The core issue is that NIR spectroscopic datasets are rarely composed of independent and identically distributed (i.i.d.) samples. For example, fruits harvested from the same orchard within the same week may share latent structure (e.g., maturity stage, temperature history, and scattering geometry), spectra acquired on the same instrument in the same session may share instrument-response characteristics. Also, in bioprocess or inline-monitoring, spectra are often collected sequentially from evolving systems, so adjacent spectra can share batch history, reactor state, media composition, operator decisions, and sampling or acquisition-protocol signatures \citep{yeung1999bioprocess,Banerjee2024FermentationICNN,Gangwar2025CellCultureCNN}. When such correlated samples are allocated to training and test sets via random splitting (a common default), performance estimates can be inflated. Under this method, the model is tested on data that share systematic structure with the training set, and accuracy metrics partly reflect this shared structure rather than genuine predictive ability under new, unseen conditions \citep{cawley2010selectionbias, roberts2017crossvalidation}. This problem arises because random splits implicitly evaluate interpolation within the sampled acquisition domain, assuming that training and test samples are approximately exchangeable. In many chemometric applications, however, the relevant deployment question is not only whether the model predicts new samples from the same acquisition distribution, but whether it remains valid under structured shifts in instrument, batch, season, temperature, scattering, sample presentation, reactor state, acquisition protocol, or process trajectory \citep{quinonero2008datasetshift}.}

\rev{Practically, this means that model performance obtained under random splits may not persist under more realistic evaluations, including instrument-transfer, seasonal-transfer, temporally blocked, batch-blocked, or process-run-blocked assessments. A few representative examples help illustrate this point.} \citet{mishra2021deepchem} showed that a CNN trained and validated in-domain suffered substantial degradation when transferred to a different portable spectrometer. \citet{walsh2024mango} found that preprocessing strategies that appeared unnecessary under in-domain evaluation became critical for maintaining accuracy under seasonal shift. \citet{dirks2022hpo} demonstrated that automatic hyperparameter optimization, when coupled to a shift-aware validation protocol (e.g., temporally blocked cross-validation), selected architectures that differed from those favored by random splits. In such settings, simpler chemometric baselines can sometimes outperform deeper CNNs, not because CNNs are inherently inferior, but because deeper models may be optimized to exploit in-domain structure that does not generalize.

\rev{This observation has a broader implication: much of the current debate about CNN architecture in chemometrics may be confounded by validation design. When Paper~A reports that a ResNet, i.e., a CNN architecture with skip connections, outperforms PLS and Paper~B reports the opposite, the divergence may reflect residual connections, but it may also reflect whether the test set shares orchard, season, instrument, temperature, batch, or operator effects with the training set. Therefore, until the field adopts structured validation protocols as a \textit{reporting standard}, architecture comparisons across studies will remain difficult to reconcile.}

\rev{Finally, validation design interacts strongly with hyperparameter tuning. Architecture, learning rate, optimizer, batch size, normalization layer, regularization strength (dropout, weight decay, early stopping), data augmentation, and preprocessing selection should be tuned against a validation objective that represents the intended deployment conditions \citep{bergstra2012randomsearch,kingma2014adam,srivastava2014dropout,ioffe2015batchnorm,passos2021automated,passos2022tutorial,dirks2022hpo}. If the validation set shares confounds with the training set, tuning will tend to select models that exploit those confounds, and model-selection bias can leak into final performance estimates if the same data are used for both selection and assessment \citep{cawley2010selectionbias}. Hence, automated hyperparameter optimization is not merely a convenience but a methodological necessity for fair comparison provided, critically, that the validation protocol itself is shift-aware and aligned with the real deployment scenario. This principle can also be exploited to build CNNs with improved robustness for a given task. An example of this is \citep{passos2023tuttifrutti} where the authors show that using spectra from different fruits to tune the hyperparameters of a CNNs for dry matter prediction, leads to models that can identify common informative bands (across species) leading to increased robustness. In practice, however, validation-aware tuning is computationally demanding: CNNs and other high-capacity neural architectures may require substantial training time even for small spectral datasets once preprocessing choices, architectural variants, random seeds, and blocked validation schemes are included in the search. Computational budget therefore becomes part of the experimental design and should be reported alongside the search space, validation protocol, and number of independent runs. The distinction between ``tuning on a random hold-out'' and ``tuning on a seasonally blocked, batch-blocked, temperature-blocked, or instrument-blocked hold-out'' is not a technical nuance; it can be the difference between selecting a robust model and selecting an overfit one. }

\section{Mapping the Contradictions}

\rev{Table~\ref{tab:study_synthesis} summarizes representative CNN and adjacent spectral-DL studies through the lens of the proposed conditional framework. It does not aim to rank models by performance, instead, it maps each study onto key moderators (e.g. domain, data regime, architecture/RF implications, validation, preprocessing, etc.) that help explain why apparently conflicting CNN-design conclusions can coexist. The table makes explicit which study features are needed to interpret apparent contradictions. Table~\ref{tab:conflicts} then summarizes the major CNN-design contradictions currently present in the Vis--NIR chemometrics literature. In it we organize each contested choice, the evidence supporting each position, and the moderating variables that can plausibly explain the disagreement (together with a proposed test that would resolve it).}

\begin{table*}[htbp]
\centering
\caption{\rev{Representative CNN and adjacent spectral-DL studies organized by moderators relevant to the conditional framework. ``NR/H'' denotes information that is not reported in a harmonized way across studies, limiting direct cross-paper comparison.}}
\label{tab:study_synthesis}
\begingroup
%\color{red}
\scriptsize
\setlength{\tabcolsep}{2pt}
\renewcommand{\arraystretch}{1.15}
\begin{tabularx}{\textwidth}{@{}>{\hsize=.52\hsize\linewidth=\hsize}Y>{\hsize=.75\hsize\linewidth=\hsize}Y>{\hsize=.75\hsize\linewidth=\hsize}Y>{\hsize=.9\hsize\linewidth=\hsize}Y>{\hsize=.95\hsize\linewidth=\hsize}Y>{\hsize=.85\hsize\linewidth=\hsize}Y@{}}
\toprule
Study & Matrix/domain and task & Data regime and availability & Architecture / ERF implication & Validation and preprocessing & Moderator lesson \\
\midrule
\citep{acquarelli2017cnn} & Vibrational spectroscopy; classification/regression examples & Benchmark-style spectral sets; sample size and resolution vary by data set & Early CNN use for vibrational spectra; learned filters replace hand-crafted features & Mainly in-domain comparisons with conventional chemometrics & Demonstrates feasibility of end-to-end spectral CNNs, but deployment shift is not the central test. \\
\citep{malek2018onecnn} & Spectroscopic signal regression & Multiple regression benchmarks; NR/H descriptors across tasks & Compact 1D-CNN baseline & Conventional split-based benchmarking & Supports parameter-efficient CNNs, especially when data are limited and tasks are benchmark-like. \\
\citep{cuifearn2018cnn} & NIR calibration; multivariate regression & Practical NIR calibration data; full reuse depends on data/split access & Modern practical CNNs with compact kernels & Compared with PLS-style calibration practice & Shows that small CNNs can be competitive, but reporting of split and tuning choices remains central. \\
\citep{zhang2019deepspectra} and \citep{Zhang2020UnderstandingCNNs} & Quantitative spectral analysis and filter interpretation & Application data with reported spectra; public reuse is limited by dataset availability & End-to-end/multi-scale CNNs; first-layer filters resemble smoothing/derivative operations & Raw-input modeling and learned preprocessing analysis & Supports learned preprocessing, but also shows why filter interpretation should be tested rather than assumed. \\
\citep{helin2021preprocessing} and \citep{bjerrum2017augmentation} & Spectral preprocessing and augmentation; pharmaceutical/NIR-style examples & Limited spectral data; augmentation and preprocessing explicitly studied & CNNs can learn or benefit from preprocessing-like transforms & EMSC/preproc./augmentation compared factorially & Raw versus preprocessed is not binary; preprocessing is a design variable. \\
\citep{ng2019soilcnn}, \citep{tsakiridis2020localizedcnn}, and \citep{yang2020cnnrnnsoil} & Soil Vis--NIR/MIR property prediction & Larger but heterogeneous soil spectral sets; data availability varies & CNN and CNN--RNN variants for multi-property regression & Cross-validation or localized validation depending on study & Evidence from soil spectra that matrix heterogeneity, locality, and validation design remain central outside food-related spectroscopy. \\
\citep{mishra2021deepchem}, \citep{dirks2022hpo} and \citep{walsh2024mango} & Mango Vis--NIR extrapolation & Application-specific fruit data; protocol details emphasized & HPO changes preferred network/training settings & Extrapolation-aware validation rather than random-only evaluation & Tuning objective and split design can change model selection. \\
\citep{gan2023dilated} and \citep{tan2023inceptionresnet} & NIR quantitative analysis and transferability & Application data; shift/transfer considered explicitly & Dilated or inception-residual designs expand effective scale & Quantitative comparison under transfer or broader validation & Large ERF can be obtained without simply increasing nominal kernel size. \\
\citep{haffner2025ipa} & Petroleum analysis & Independent industrial matrix; partial data availability & Inception-based CNN for chemicals spectral analysis & Application-oriented validation & Broadens evidence beyond biological samples while preserving the same scale/protocol questions. \\
\bottomrule
\end{tabularx}
\endgroup
\end{table*}

\begin{table*}[htbp]
\centering
\caption{Examples of major CNN design contradictions in Vis-NIR chemometrics and likely moderators.}
\label{tab:conflicts}
\begingroup
\scriptsize
\setlength{\tabcolsep}{3pt}
\renewcommand{\arraystretch}{0.98}
\begin{tabularx}{\textwidth}{@{}>{\hsize=.65\hsize\linewidth=\hsize}Y>{\hsize=1.05\hsize\linewidth=\hsize}Y>{\hsize=1.05\hsize\linewidth=\hsize}Y>{\hsize=1.25\hsize\linewidth=\hsize}Y@{}}
\toprule
Contested choice & Evidence supporting option A & Evidence supporting option B & Likely moderator(s) and decisive test \\
\midrule
Small versus large kernels & Small/compact CNNs can be competitive and parameter-efficient \citep{malek2018onecnn,cuifearn2018cnn,chenwang2019e2e} & Multi-scale, residual, or dilated designs outperform in broader or shifted settings \citep{zhang2019deepspectra,gan2023dilated,martins2022spectranet53,tan2023inceptionresnet} & Feature width relative to kernel ERF; match parameter budget and ERF across models; compare under in-domain and external splits. \\

Classification versus regression & Classification can reward local discriminative cues and compact representations \citep{passos2021automated,zhang2022mlreview} & Quantitative regression for broad latent traits benefits from larger ERF and multi-scale context \citep{cuifearn2018cnn,gan2023dilated,martins2023spectranet32,mishra2022multioutput} & Keep data and preprocessing fixed; compare task heads and ERF profiles across task families. \\

Shallow versus deeper models & Simpler models avoid overfitting and are easier to tune with limited data \citep{cuifearn2018cnn,dirks2022hpo,einarson2021pectin} & Deeper residual/multi-branch models improve accuracy when data coverage is adequate \citep{he2016resnet,martins2022spectranet53,passos2023tuttifrutti} & Sample size, augmentation, and regularization determine whether depth helps; run depth sweeps at fixed training protocol. \\

\rev{Architecture-only comparisons} & \rev{Kernel size and depth are often reported as the main design variables \citep{malek2018onecnn,gan2023dilated}} & \rev{Optimizer, batch normalization, batch size, early stopping, dropout/weight decay, and learning-rate schedule can also change reproducibility \citep{kingma2014adam,srivastava2014dropout,ioffe2015batchnorm,dirks2022hpo}} & \rev{Report and tune the full training protocol; compare architectures only after controlling non-kernel hyperparameters.} \\

Raw spectra versus preprocessing & End-to-end raw modeling can match or exceed conventional pipelines \citep{chenwang2019e2e,zhang2019deepspectra} & Preprocessing and augmentation improve generalization, especially under shift \citep{helin2021preprocessing,mishra2021synergy,passos2022tutorial,walsh2024mango} & Treat preprocessing as a factorial variable, not a fixed pre-step; report interactions with architecture. \\

Single-domain training versus transfer learning & Direct model reuse may fail under season/instrument shift \citep{mishra2021deepchem,guo2023transfer} & Fine-tuning and transfer-specific architectures improve adaptation \citep{mishra2021caltransfer,yang2022dts,tan2023inceptionresnet,yang2022interseason} & Evaluate transfer with controlled target-label budgets (zero-shot, few-shot, full fine-tune). \\

\rev{Closed or weakly described data versus reusable data/splits} & \rev{Application papers often report only aggregate metrics and leave sample provenance, splits, or code unavailable} & \rev{Reusable data, fixed splits, and code enable independent tests of ERF, preprocessing, validation design, and architecture as performance moderators \citep{wilkinson2016fair,pineau2020reproducibility,luo2017erf,dirks2022hpo}}  & \rev{Treat data/code availability as an evidential weight, not as an administrative afterthought.} \\

Accuracy-only versus interpretable modeling & High predictive scores alone are common but mechanism remains uncertain \citep{jia2024aidriven,bec2025interpretability} & Interpretable variable attribution can identify chemically plausible spectral regions \citep{duan2023perturbator,akulich2022xai,yang2022dts} & Require faithfulness tests (ablation/insertion, seed stability, randomization) alongside saliency visuals. \\
\bottomrule
\end{tabularx}
\endgroup
\end{table*}

The ``small versus large kernels'' conflict is arguably the most frequently encountered and, simultaneously, the most poorly framed. As argued in the preceding sections, this is fundamentally a question about ERF relative to informative feature width. %When informative features are narrow, small kernels operating as local derivative-like filters can be sufficient and parameter-efficient. Conversely, when informative features are broad, as in SSC regression mediated by water-band deformation, large ERFs should work better, whether achieved via large kernels, dilation, adequate depth, or explicit multi-scale processing.
Therefore, studies that compare ``small'' and ``large'' kernels without equalizing ERF and parameter budget are effectively testing a confounded comparison. Although their conclusions may be internally valid for that specific setup, they are difficult to generalize.

The ``raw versus preprocessing'' debate deserves particular attention because it is often discussed as a matter of principle, rather than as a question of how preprocessing and the learned model interact in practice. If CNNs can learn derivative-like operations in their early layers \citep{cuifearn2018cnn,helin2021preprocessing,passos2025deeptuttifrutti2}, explicit preprocessing may appear redundant. However, this argument underestimates two practical realities. First, learning a robust preprocessing surrogate from data requires sufficient training coverage, and the small datasets typical of NIR chemometrics may not supply enough signal for the network to reliably learn the optimal transform jointly with the prediction task. Second, certain preprocessing operations such as multiplicative scatter correction (MSC) or standard normal variate (SNV) normalization, target physical effects (e.g., particle size and scattering) that are largely orthogonal to the chemistry labels. From a machine learning point of view this means that the neural network has little incentive to learn them reliably from chemistry-supervised loss alone. Hence, what we propose as a pro-active solution is not to choose sides, but to treat preprocessing as a co-designed (and co-optimized) component of the overall model design space \citep{mishra2021synergy,passos2022tutorial} and include it explicitly in hyperparameter optimization. \rev{This, of course, is no novelty because optimizing preprocessing together with the model (usually non-DL) is part of the chemometricians' playbook.}

The transfer-learning conflict highlights a practical bottleneck that extends beyond architecture. Calibration transfer across instruments, seasons, and production conditions has been central to NIR spectroscopy long before deep learning \citep{roger2003epo}. Deep learning introduces powerful tools (e.g. fine-tuning pretrained representations, domain-adversarial training, or self-supervised pretraining) but it also introduces new failure modes, since deep models can encode instrument-specific signatures efficiently \citep{mishra2021deepchem,mishra2021caltransfer,guo2023transfer}. Consequently, the key methodological requirement is to evaluate transfer under controlled target-label budgets, i.e., how many labeled samples from the new domain are needed to recover acceptable accuracy (zero-shot, few-shot, and full fine-tuning). Without this information, claims about ``transferable'' architectures remain difficult to substantiate. An additional and potentially complementary direction is to design models with explicit mechanisms for in-context adaptation or test-time compute \citep{olsson2022inductionheads,brown2020fewshot}, i.e., models that can adjust their predictions using a small set of target-domain samples at inference time rather than relying solely on offline fine-tuning. Although this paradigm remains largely unexplored in NIR chemometrics, it could be particularly impactful for calibration transfer, where rapid adaptation to new instruments or operating conditions is often required.

\rev{A final cross-cutting limitation is reproducibility. Many CNN--NIR studies rely on proprietary or application-specific datasets and report only aggregate performance metrics, while the exact sample identifiers, split seeds, acquisition blocks, preprocessing settings, and tuning budgets are not always made available. As a result, it can be difficult to determine whether a reported difference reflects an architectural mechanism, the composition of the split, or unreported correlations in the data. This also prevents independent reanalysis under matched RF/parameter budgets, alternative preprocessing choices, or shift-aware validation protocols. Thus, data availability, code availability, and split reporting should be treated as part of the evidential weight of CNN--NIR comparisons rather than as administrative details. When raw spectra cannot be released, authors should at minimum report sample provenance and blocking variables and provide fixed train/validation/test indices, or an unambiguous split-generation rule, so that architectural claims can be meaningfully reproduced and stress-tested \citep{jia2024aidriven,luo2024cnnreview}.}

\section{\rev{CNNs and Transformer-Based Spectral Models}}

\rev{Although the present review focuses on CNNs, Transformer-based and hybrid CNN--Transformer models are increasingly relevant for spectral analysis. The conceptual contrast is straightforward. CNNs impose a locality prior: nearby wavelengths are processed together first, and larger-scale context emerges through larger kernels, depth, dilation, pooling, or multi-scale branches. Transformers replace this fixed local prior with attention over tokenized inputs, allowing distant wavelength regions to interact directly \citep{vaswani2017attention,dosovitskiy2021vit}. This is attractive for spectroscopy because chemically related features can be separated in wavelength, and because broad matrix effects may require comparing non-adjacent regions. Hyperspectral Transformer models such as SpectralFormer explicitly exploit spectral sequence structure and have shown that attention-based models can capture band-to-band dependencies in high-dimensional spectral data \citep{hong2021spectralformer}.
Given the relevance that attention has in Transformers, some works have explored (with different degrees of success) CNNs extended with different attention mechanisms \citep{Zou2021SEResNet,ZHANG2025_CNNAttention,Li2024_CNNAttention, LIU2026_CNNAttention}.}

\rev{However, the advantages of attention do not remove the moderators emphasized above. First, Transformers are often more data-hungry than compact CNNs unless strong priors, pretraining, or hybrid tokenizers are used; this is a serious constraint for many chemometric datasets \citep{han2023vitreview}. Because standard Transformers do not inherently encode the local weight sharing and translation equivariance built into convolutions, they provide a weaker prior for detecting local slopes, shoulders, and band-shape patterns along the wavelength axis. Unless locality is introduced through convolutional tokenizers, spectral patches, relative positional encodings, pretraining, or other architectural priors, these relationships must be learned more directly from the data, increasing sample-size requirements and overfitting risk in small chemometric datasets. Second, attention maps are not automatically mechanistic explanations: they identify learned interactions between tokens, but still require faithfulness, stability, and randomization checks before being interpreted chemically \citep{jain2019attention}. Third, robustness under instrument, temperature, batch, or season shift remains an empirical question. A hybrid CNN--Transformer model may be useful when the CNN front-end captures local spectral shape while attention layers model longer-range dependencies, consistent with convolutional-tokenizer designs in the vision literature \citep{wu2021cvt}, but it should be compared against CNN and PLS baselines under the same split, tuning budget, preprocessing search space, and target-domain label budget. Thus, although Transformers broaden the architecture space, they do not provide a universal design recipe yet; they inherit the same need for physics-aware priors and deployment-aligned validation.}

\section{A Conditional Design Framework}

\rev{The evidence reviewed above supports a single organizing principle: \textit{architecture choices in CNN-based Vis--NIR chemometrics are conditional on the interaction of spectral physics, data regime, acquisition protocol, and intended deployment scenario}. A single ``best'' architecture may exist in principle, but the current literature does not support a universal recipe for CNN design. Below we propose a framework or decision workflow that could guide the user to find a suitable architecture for specific datasets, not a fixed architecture prescription. It should be based on:}

\rev{\textit{Element~1: kernel scale as a physics-aligned prior.} Kernel design should be linked to the expected width of the informative spectral structure. A practical starting rule is to express kernel size in wavelength-sample units relative to the expected band deformation width. If an informative deformation spans $w$ spectral points (e.g., an 80~nm water-band deformation at 2~nm resolution corresponds to $w\approx 40$ points), then initial kernels in the approximate range $w/8$ to $w/2$ (with a lower bound of 3) provide a more defensible starting prior than fixed values borrowed from computer vision heuristic rules. This range should not be treated as a recipe: it is a search-space anchor to be stress-tested. Importantly, it should be cross-checked against the resulting network-level ERF to ensure the model can, in principle, ``see'' the full feature extent. Multi-scale modules and dilation provide robust compromises because they reduce dependence on any single manually chosen kernel scale \citep{szegedy2016inception,zhang2019deepspectra,gan2023dilated,yu2023multiscale}.}

\rev{\textit{Element~2: depth, capacity, and the small-data regime.} For the small-to-moderate datasets that dominate chemometric practice (often $n<1000$), deeper models trade representational capacity for increased overfitting risk and tuning sensitivity. Therefore, automated hyperparameter search over depth, channel width, learning rate, optimizer, batch size, batch normalization, dropout, weight decay, early stopping, and learning-rate schedule is not a luxury but a methodological necessity for rigorous model comparison \citep{kingma2014adam,srivastava2014dropout,ioffe2015batchnorm,passos2021automated,passos2022tutorial,dirks2022hpo}. The critical requirement is that the validation objective used during tuning must represent the target deployment condition: tuning against a random in-domain split tends to select models that exploit within-domain correlations, whereas tuning against a seasonally blocked, batch-blocked, temperature-blocked, or instrument-blocked split tends to select models that are robust to the relevant sources of shift.}

\rev{\textit{Element~3: preprocessing as a co-optimized categorical factor.} Treating ``raw versus preprocessed'' as a binary choice seems a false dilemma. Preprocessing type (none, SNV, MSC, EMSC, first derivative, second derivative, smoothing, baseline correction, and combinations thereof) should be included as a categorical hyperparameter within the joint optimization space, alongside architecture and training parameters \citep{bjerrum2017augmentation,helin2021preprocessing,mishra2021synergy}. This enables automated search to discover preprocessing--architecture interactions rather than fixing one component before optimizing the other. The observation that CNNs can learn derivative-like first-layer filters does not eliminate the value of explicit preprocessing; it implies that the optimal pipeline is data-dependent and should be selected empirically.}

\rev{\textit{Element~4: acquisition and transferability as first-order objectives.} In real deployment, a model will inevitably encounter instruments, seasons, lots, temperatures, operators, matrices, or environmental conditions not represented in the training data. Therefore, acquisition protocol and transfer-ready pipelines (e.g. fine-tuning strategies or domain-adaptation techniques) should be designed into the workflow from the outset, rather than retrofitted after in-domain performance has been optimized \citep{roger2003epo,liu2017temperature,mishra2021caltransfer,yang2022dts,tan2023inceptionresnet}. A minimum reporting standard for transfer claims should include performance under zero-shot (no target labels), few-shot (10--50 target labels), and full fine-tuning conditions, reported separately to make the practical adaptation cost explicit.}

\rev{Operationally, the workflow can be summarized as follows: (1) identify the expected spectral feature width and main confounds (temperature, scattering, instrument, batch, season); (2) translate feature width into an initial ERF/kernel/dilation/multi-scale search space; (3) include preprocessing and non-kernel training choices in the same optimization budget; (4) choose validation splits that block the deployment-relevant confound; (5) report data/code availability, uncertainty across seeds, and target-domain label budget; and (6) interpret model attributions only after faithfulness and stability tests. These steps are deliberately conditional. They help researchers design a defensible comparison, but they cannot replace empirical validation on the target matrix.}

\section{Interpretability as Validation Protocol}

Interpretability in CNN-based NIR chemometrics operates at two coupled levels: (i)~\textit{spectral--chemistry interpretation} (which wavelengths or spectral regions contribute to the prediction, and do they align with known molecular assignments?), and (ii)~\textit{model--mechanism interpretation} (how does the network's internal computation produce the prediction?) \citep{bec2025interpretability}. Most published work addresses primarily the first level, typically by producing saliency maps, attention-weight visualizations, \rev{SHAP/LIME-type local explanations, or perturbation profiles} and then checking whether highlighted wavelengths overlap with known absorption bands \citep{ribeiro2016lime,lundberg2017shap,duan2023perturbator,akulich2022xai,yang2022dts,kim2024citrus,passos2025deeptuttifrutti2}. \rev{The second level is less often addressed, but it can be probed by analyzing learned first-layer filters, intermediate feature maps, channel activations, or the effect of removing branches/filters that are hypothesized to encode specific spectral scales \citep{Zhang2020UnderstandingCNNs}.}

While intuitive, this practice faces a central limitation: plausibility is not evidence of mechanism. \rev{We fall back again to fruit spectra as an illustrative example}. A saliency map that highlights the 970~nm water band for SSC prediction is consistent with an aquaphotomics-mediated mechanism, but it is also consistent with confounds (e.g., temperature-driven water shifts or scattering artifacts) that correlate with SSC within the training domain. \rev{Furthermore, popular additive explanation methods such as SHAP or LIME assume local feature independence. This assumption is heavily strained by the extreme collinearity of spectral data, where adjacent wavelengths are physically coupled by broad molecular vibrations; consequently, point-wise attributions can be dispersed, unstable, or misleadingly concentrated on noise artifacts if collinearity is not explicitly addressed \citep{zhang2026shapca}.} Without additional tests, plausible attribution is difficult to distinguish from confound-mediated attribution. \rev{Table~\ref{tab:xai_methods} summarizes common explanation families and the main caveats in spectral applications.}

\begin{table*}[htbp]
\centering
\caption{\rev{Common explainability methods for spectral CNNs and limitations that should be reported before mechanistic interpretation.}}
\label{tab:xai_methods}
\begingroup
%\color{red}
\small
\setlength{\tabcolsep}{3pt}
\renewcommand{\arraystretch}{1.12}
\begin{tabularx}{\textwidth}{@{}>{\hsize=.75\hsize\linewidth=\hsize}Y>{\hsize=1.05\hsize\linewidth=\hsize}Y>{\hsize=1.25\hsize\linewidth=\hsize}Y>{\hsize=1.05\hsize\linewidth=\hsize}Y@{}}
\toprule
Method family & What it reports & Main spectral caveat & Minimum check \\
\midrule
Occlusion / perturbation / ablation & Performance or prediction change when wavelengths or regions are removed, permuted, or replaced & Correlated wavelengths make single-channel removal unstable; window size can determine the result & Report region-wise ablation, insertion/deletion curves, and sensitivity to window width \citep{duan2023perturbator}. \\
Gradient, saliency, integrated gradients, Grad-CAM variants & Local derivative or activation sensitivity around a trained model & Gradients can be noisy, saturated, or insensitive to data randomization \citep{sundararajan2017ig,selvaraju2017gradcam,adebayo2018sanity} & Repeat across seeds; include randomized-label or randomized-weight sanity checks. \\
SHAP / LIME-type local surrogate explanations & Approximate feature contributions for a prediction & Feature independence assumptions are strained by highly collinear spectra \citep{ribeiro2016lime,lundberg2017shap} & Use grouped wavelength regions and compare with perturbation tests. \\
Attention maps and attention layer weights & Learned token-to-token interactions & Attention weights are not automatically causal explanations & Validate by masking high-attention regions and testing prediction/metric degradation. \\
Filter and feature-map probing & Internal model-mechanism evidence: derivative-like filters, branch specialization, or scale selectivity & Visual resemblance to known preprocessing is suggestive but not sufficient & Quantify filter/branch ablations and compare learned filters across seeds \citep{Zhang2020UnderstandingCNNs}. \\
\bottomrule
\end{tabularx}
\endgroup
\end{table*}

We therefore argue that interpretability in this field should function as a \textit{falsification protocol} rather than as a post-hoc narrative. At minimum, interpretability analyses should include the following operational checks:
\rev{\begin{enumerate}
\item \textit{Faithfulness:} quantify the performance or prediction drop when highlighted regions are removed, permuted, occluded, or reinserted.
\item \textit{Stability:} repeat attributions across random seeds, bootstrap resamples, and plausible preprocessing variants.
\item \textit{Sanity checks:} compute attributions for randomized labels or randomized weights to ensure the explanation method is not generating spurious spectral structure \citep{adebayo2018sanity}.
\item \textit{Method agreement:} compare perturbation-based, gradient-based, decomposition-based, and filter/feature-map analyses to identify regions consistently highlighted across methodologies \citep{duan2023perturbator,bec2025interpretability}.
\end{enumerate}}
These quantitative checks should be complemented by comparison against spectroscopic priors (e.g., water-sensitive regions, analyte combination bands, and pigment features) to assess whether attributions are physically meaningful.

For water-dominated traits such as SSC, the aquaphotomics perspective suggests a concrete, pre-registered interpretability criterion: \textit{shoulder sensitivity}. If prediction is genuinely driven by solute-induced deformation of a broad water band, attribution should concentrate on the flanks (i.e., shoulders) rather than at the peak maximum, because intensity changes induced by band shifts or FWHM changes are maximized near the steepest slopes of the absorption profile. \rev{Pre-registration in this context would mean specifying, before model training or explanation inspection, the shoulder-region wavelength bounds, the perturbation window, the expected direction of metric degradation, and the minimum degradation threshold required to support the proposed mechanism.} This prediction can be tested via targeted shoulder-region ablation experiments, providing a quantitative and falsifiable test of the measurement mechanism rather than a purely qualitative visual assessment.

\rev{Compared to spectral–chemistry interpretation, \textit{mechanistic interpretability} remains mostly underdeveloped for CNN--NIR chemometrics. In this setting, mechanism should not mean only that an attribution profile overlaps a chemically plausible band; it should mean that a proposed computational pathway can be tested from spectral perturbation to internal representation and, finally, to prediction. Existing work provides partial examples of this direction: convolutional filters have been interpreted as learned spectral transformations or preprocessing operations \citep{acquarelli2017cnn,bjerrum2017augmentation,cuifearn2018cnn,Zhang2020UnderstandingCNNs}, attribution methods have been used to compare CNN reliance with PLS/VIP-like spectral importance and known vibrational assignments \citep{duan2023perturbator,passos2025deeptuttifrutti2,bec2025interpretability}, and perturbation-based tests can quantify whether highlighted regions are actually necessary for prediction. However, true mechanism-level evidence remains uncommon because many analyses stop at visually plausible saliency maps or filter shapes. Neural network mechanistic interpretability is particularly challenging in NIR chemometrics because the role learned by a convolutional block may change with the dataset distribution, preprocessing, chemometric target, sample matrix, and acquisition conditions, and because the field lacks a standardized NIR benchmark suite for testing whether such roles generalize across tasks.}

\section{Discussion}

The central point of this review is that many hyperparameter contradictions in CNN--NIR papers are not evidence of a chaotic field; rather, they are evidence of \textit{incomplete conditioning}. \rev{Architecture performance depends on the interaction of spectral physics, data regime, acquisition protocol, and validation design, and when these moderating variables are not controlled, conflicting results are an expected outcome.}

This perspective has immediate practical consequences. The question ``what is the best CNN architecture for NIR spectroscopy?'' is ill-posed in the same way that ``what is the best statistical model?'' is ill-posed without specifying the data-generating process, sample size, and evaluation criterion. \rev{A more productive formulation is: ``given the spectral feature width, dataset size, acquisition protocol, expected deployment shift, and confound structure of my application, what architecture family and validation strategy should I prioritize?'' }The conditional framework and guideline matrix presented here are designed to help answer this latter question by making moderating variables explicit and linking them to actionable design choices.

Regarding the ``raw versus preprocessing'' question, the evidence supports a conditional position (rather than binary). End-to-end raw-spectrum pipelines are credible and often strong within the training domain \citep{chenwang2019e2e,zhang2019deepspectra}, however, their superiority is not universal. Under explicit shift and confound pressure, preprocessing-aware or hybrid chemometric--DL strategies can be decisively better \citep{helin2021preprocessing,mishra2021synergy,walsh2024mango}. Therefore, preprocessing should be best treated as a co-optimized hyperparameter, not as a fixed initial choice.

Depth exhibits similar conditionality. Deeper residual and multi-scale models can outperform compact models when training coverage is sufficient \citep{martins2022spectranet53,passos2023tuttifrutti}, whereas simpler models may generalize better when training coverage is narrow or validation design is weak \citep{dirks2022hpo,mishra2021deepchem}. We argue that a common failure mode is not that deep models are intrinsically worse, but that they can overfit within-domain structure more efficiently, producing apparently strong performance under random splits and poor performance under external evaluation.

On interpretability, the field faces a credibility challenge. High predictive performance is insufficient for mechanistic claims in systems dominated by indirect measurement. The practice of generating saliency maps, \rev{Grad-CAM scores, SHAP/LIME profiles, or attention maps, observing that they ``agree with known spectroscopy,'' and declaring the model interpretable falls short of the standard required for scientific inference.} Hence, the field should move from qualitative saliency narratives toward falsifiable explanation protocols anchored in spectral plausibility, perturbation-based faithfulness testing, \rev{sanity checks, and pre-specified interpretability criteria (e.g., the shoulder-sensitivity test proposed above) \citep{duan2023perturbator,adebayo2018sanity,bec2025interpretability}.}

\subsection{Practical Guidelines for Researchers}

Table~\ref{tab:guidelines} translates the synthesis above into a scenario-specific "decision matrix" that can help users design  CNN-based Vis-NIR models (i.e., mapping common application regimes to recommended model families, tuning priorities, and minimum validation/interpretation protocols).

\begin{table*}[htbp]
\centering
\caption{Conditional guidelines for CNN-based Vis-NIR chemometrics.}
\label{tab:guidelines}
\begingroup
\small
\setlength{\tabcolsep}{3pt}
\renewcommand{\arraystretch}{1.12}
\begin{tabularx}{\textwidth}{@{}>{\hsize=.90\hsize\linewidth=\hsize}Y>{\hsize=1.05\hsize\linewidth=\hsize}Y>{\hsize=.95\hsize\linewidth=\hsize}Y>{\hsize=1.10\hsize\linewidth=\hsize}Y@{}}
\toprule
Scenario & Recommended model family & Hyperparameter priorities & Minimum validation and interpretation protocol \\
\midrule
Limited data ($n<1000$), moderate shift risk & Compact CNN or shallow residual with strong regularization \citep{cuifearn2018cnn,dirks2022hpo,einarson2021pectin} & Conservative depth, dropout/L2, simple kernels plus small dilation sweep & Blocked split by batch/season; report seed variability and confidence intervals. \\

Medium data ($10^3$--$10^4$), broad bands and mixed scales & Multi-scale CNN (Inception-like) or residual multi-branch \citep{zhang2019deepspectra,martins2022spectranet53,yu2023multiscale} & Joint tuning of kernel scales and ERF; early stopping on shift-aware validation & Include ablation for branch removal and kernel-scale sensitivity. \\

High shift risk (instrument/season/orchard) & Transfer-ready pipelines with fine-tuning and domain adaptation \citep{mishra2021caltransfer,yang2022dts,tan2023inceptionresnet} & Fine-tune depth and layer-freeze policy; optimize with target-domain validation & Report zero-shot, few-shot, and full adaptation performance separately. \\

Strong confound risk (temperature, maturity, color) & Physics-aware hybrid strategy (chemometric correction plus DL) \citep{sun2020temp,mishra2021synergy,roger2022aquapre} & Preprocessing-architecture factorial search; explicit confound augmentation & Perform stress tests where confound--target correlation is perturbed. \\

\rev{Large reusable data or explicit long-range dependencies} & \rev{CNN+attention baseline or hybrid CNN--Transformer or spectral Transformer comparison \citep{hong2021spectralformer,chen2024spectraformer,jin2023swintransformer}} & \rev{Tokenization scale, positional encoding, pretraining, attention dropout, and CNN front-end width} & \rev{Compare against compact CNNs at matched tuning budget; include attention faithfulness tests and external-domain splits.} \\

Interpretability-critical applications & Architectures with perturbation-based variable attribution \citep{duan2023perturbator,bec2025interpretability} & Favor stable models over marginal score gains; enforce attribution stability & Add faithfulness, stability, and sanity checks before mechanistic claims. \\
\bottomrule
\end{tabularx}
\endgroup
\end{table*}

In addition to the scenario-specific guidance, we recommend the following \textit{minimum reporting protocol} for all future CNN--NIR publications (regardless of application), because incomplete reporting is a major driver of irreproducible comparisons:
\begin{enumerate}
\item \rev{\textit{Complete architecture and training specification:} kernel sizes, strides, dilation factors, activation functions, normalization layers, optimizer, learning-rate schedule, batch size, regularization (dropout rate, weight decay), early stopping rule, and total parameter count.}
\item \textit{Full split logic:} rationale for the split strategy (random, blocked, external), identity of blocking variables, random seed(s) used, and the number of repeated runs.
\item \textit{Preprocessing and augmentation:} complete pipeline specification including spectral range selection, preprocessing type and parameters, and any data augmentation applied during training, reported as reproducible code or pseudocode.
\item \rev{\textit{Data and code availability:} public data/splits/code where possible, following FAIR principles where applicable \citep{wilkinson2016fair}; otherwise, a complete description of sample provenance, acquisition protocol, split identifiers, and reuse limitations.}
\item \textit{Uncertainty reporting:} bootstrap confidence intervals, prediction intervals, or repeated-run standard deviations for all reported metrics.
\item \textit{Interpretation protocol:} quantitative faithfulness tests (performance drop on feature ablation), attribution stability across seeds, and sanity checks against randomized controls where mechanistic claims are made.
\end{enumerate}

\section{Conclusions}

\rev{In principle, a universally superior CNN (or other type of neural network) architecture for Vis--NIR chemometrics, i.e. a sort of spectral foundation model, may be created in the future and overcome most of the mentioned hurdles. However, the present evidence }reviewed here suggests that \textit{current} disagreements are driven less by irreconcilable methods than by incomplete conditioning on spectral physics, data regime, and validation design. Accordingly, rather than promoting a single ``best'' template, we argue that the most \rev{useful} near-term research strategy is a conditional design framework organized around four practical principles:
\begin{enumerate}
\item \textit{Physics-aligned scale:} Link architectural scale (kernel size, depth, dilation, and multi-scale structure) to the physical scale of informative spectral variation, recognizing that the effective receptive field (ERF), not nominal kernel width, is the operative quantity.
\item \rev{\textit{Deployment-aligned validation:} Tune and compare models under validation protocols that reflect the intended deployment shift (e.g., season, instrument, batch, temperature), since split design can rival or exceed architecture effects in some applications.}
\item \textit{Preprocessing as a design variable:} Treat preprocessing and augmentation as co-designed components of the pipeline, jointly optimized with architecture rather than fixed upstream by convention.
\item \textit{Robustness and explanation as requirements:} Treat transfer performance (zero-shot, few-shot, and full fine-tuning) and interpretability stress tests (faithfulness, stability, and sanity checks) as prerequisites for deployment claims, not optional add-ons.
\end{enumerate}

\rev{Taken together, these principles reframe CNN development for Vis--NIR spectral analysis from narrative-driven architecture selection toward controlled, falsifiable comparisons. In practical terms, progress now depends less on proposing ever more complex networks and more on adopting shared benchmarks, minimum reporting standards, and factorial stress tests that isolate moderators (ERF, preprocessing, tuning budget, acquisition protocol, data availability, and shift exposure). Such a strategy would allow the field to converge (either toward task-specific families of architectures or, potentially, toward a genuinely robust foundation model) with conclusions supported by reproducible evidence rather than by split-dependent rankings.}

\section*{Acknowledgements}
The author thanks the SensAIfood CIG (IG19145) for providing collaboration network support. This review is based on class materials presented by the author in the SensAIFood "Training School in AI methods applied to spectral data", that took place at CRA-W in Namur, Belgium, May 2025 and in the SensAIfood Final Conference, in Tirana, Albania in October 2025.

\section*{Funding}
D. Passos acknowledges funding by FCT/RNCA projects 2024.10078.CPCA.A1 and 2025.12264.CPCA.A1. The funders had no role in the preparation of the manuscript or in the decision to submit it for publication.

\section*{Declaration of generative AI and AI-assisted technologies in the manuscript preparation process}
During the preparation of this work, the author used Gemini-3.1 (Google) and ChatGPT-5.2 (OpenAI) to refine the language and improve the coherence and clarity of the manuscript. After using these tools, the author reviewed and edited the content as needed and takes full responsibility for the content of the publication.

\section*{CRediT authorship contribution statement}
Dário Passos: Conceptualization, Formal analysis, Investigation, Methodology, Visualization, Writing - original draft, Writing - review and editing.

\section*{Declaration of competing interest}
The author declares that he has no known competing financial interests or personal relationships that could have appeared to influence the work reported in this paper.

\section*{Data availability}
No new data were generated or analyzed in this review.

\bibliographystyle{elsarticle-num-names}
\bibliography{references}

\end{document}